\pdfoutput=1

\documentclass[11pt]{article}

\usepackage[]{EMNLP2022}

\usepackage{times}
\usepackage{latexsym}

\usepackage[T1]{fontenc}

\usepackage[utf8]{inputenc}

\usepackage{microtype}

\usepackage{inconsolata}
\usepackage{amsmath,graphicx}
\usepackage{todonotes,comment,pbox,multirow,arabtex,utf8,pifont,graphics,amsmath,textcomp,subcaption,array,float,adjustbox,subcaption,caption,pbox}

\newcommand{\hide}[1]{}

\newcommand{\AMADDA}{{\={A}}}

\title{ArzEn-ST: A Three-way Speech Translation Corpus for \\Code-Switched Egyptian Arabic - English}

\author{Injy Hamed,$^{1,2}$ Nizar Habash,$^{1}$ Slim Abdennadher,$^{3}$ Ngoc Thang Vu$^{2}$ \\
  $^1$Computational Approaches to Modeling Language Lab, 
  New York University Abu Dhabi \\
  $^2$Institute for Natural Language Processing, University of Stuttgart\\
  $^3$Computer Science Department, The German University in Cairo\\
  \texttt{injy.hamed@nyu.edu}
  }

\begin{document}
\maketitle
\begin{abstract}
We present our work on collecting ArzEn-ST, a code-switched Egyptian Arabic - English Speech Translation Corpus. This corpus is an extension of the ArzEn speech corpus, which was collected through informal interviews with bilingual speakers. In this work, we collect translations in both directions, monolingual Egyptian Arabic and monolingual English, forming a three-way speech translation corpus. We make the translation guidelines and corpus publicly available. We also report results for baseline systems for machine translation and speech translation tasks. We believe this is a valuable resource that can motivate and facilitate further research studying the code-switching phenomenon from a linguistic perspective and can be used to train and evaluate NLP systems.
\end{list} 
\end{abstract}

\setcode{utf8}
\section{Introduction}
Code-switching (CSW), defined as the alternation of language
in text or speech, is a common linguistic phenomenon in multilingual societies. CSW can occur on the boundaries of sentences, words (within the same sentence), or morphemes (within the same word). 
While the worldwide prevalence of CSW has been met with increasing efforts in NLP systems trying to handle such mixed input, data sparsity remains one of the main bottlenecks hindering the development of such systems \cite{CCV16}.

\begin{figure}[t]
    \centering
    \includegraphics[width=\columnwidth]{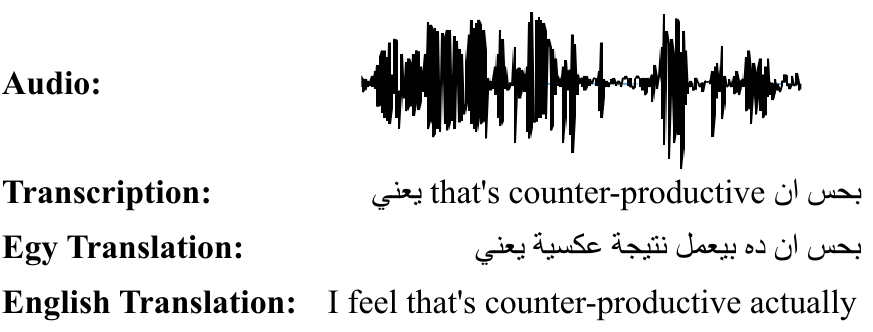}
    \caption{An example from the corpus, showing the four representations for each utterance: audio, transcription, Egyptian Arabic translation, and English translation.}
    \label{fig:corpus_example}
\end{figure}

In this paper, we present  ArzEn-ST,\footnote{Arz is the ISO 639-3 code for Egyptian Arabic.} a speech translation (ST) corpus for code-switched 
Egyptian Arabic (Egy)~-~English.
We extend the ArzEn Egyptian Arabic-English CSW conversational speech corpus \cite{HVA20} with translations going to both directions; the primary (Egyptian Arabic) as well as secondary (English) languages. See Figure~\ref{fig:corpus_example}. 
This corpus is a valuable resource filling an important gap, given the naturalness and high frequency of CSW in it. It can be used for the purpose of linguistic investigations as well as for building and evaluating NLP systems. We provide benchmark baseline results for the tasks of automatic speech recognition (ASR), machine translation (MT), and ST.
We make the translation guidelines and full corpus available, as well as the experiments' scripts and data splits.\footnote{\url{http://arzen.camel-lab.com/}\label{fnsite}}

The paper is organized as follows. In Section~\ref{sec:related_work}, we provide an overview of previous work done for code-switched ASR, MT, and ST tasks as well as corpora collection. In Section~\ref{sec:arzen}, we provide an overview of the ArzEn speech corpus. In Section~\ref{sec:translation_guidelines}, we elaborate on the translation guidelines used to create the three-way parallel ST corpus. Finally, in Section~\ref{sec:baseline_systems}, we report the performance of the ASR, MT, and ST baseline systems. 

\section{Related Work}
\label{sec:related_work}

\begin{table*}[t]
\begin{tabular}{|p{7cm}|p{8cm}|}
\hline
\multicolumn{1}{|c|}{\textbf{Language}} & \multicolumn{1}{c|}{\textbf{Citations}} \\\hline\hline
\multicolumn{2}{|c|}{\textbf{Translating CSW $\rightarrow$ monolingual}}\\\hline\hline
Hindi-English $\rightarrow$ English    &   \cite{DKS18,SS20,TKP21,CAS+22}\\
Sinhala-English $\rightarrow$ Sinhala  &   \cite{KS21}\\
English-Spanish $\rightarrow$ English  &   \cite{CAS+22}\\
MSA-Egyptian Arabic $\rightarrow$ English   &  \cite{CAS+22}\\
English-Bengali $\rightarrow$ both     &   \cite{MMD+19}\\
Egyptian Arabic -English $\rightarrow$ both    &   ArzEn-ST (the corpus presented in this paper) \\\hline\hline
\multicolumn{2}{|c|}{\textbf{Translating monolingual $\rightarrow$ CSW}}\\\hline\hline
Hindi $\rightarrow$ Hindi-English      &   \cite{TKP21,BMA+18}\\
English $\rightarrow$ Bengali-English  &    \cite{BMA+18}\\
English $\rightarrow$ Gujarati-English &   \cite{BMA+18}\\
English $\rightarrow$ Tamil-English    &   \cite{BMA+18}\\
English, Hindi $\rightarrow$ Hindi-English  &   \cite{SS21}\\\hline
\end{tabular}
\caption{Overview on available human-annotated CSW-focused parallel corpora.}
\label{table:MT_corpora_rw}
\end{table*}
\subsection{CSW Automatic Speech Recognition} CSW ASR has gained a considerable amount of research \cite{VLW+12,LV19,ACH+21,HDL+22,HCA+22}, where several CSW speech corpora have been collected, covering multiple language pairs, including Chinese-English 
\cite{LTC+15}, Hindi-English 
\cite{RS17}, Spanish-English 
\cite{SL08}, Arabic-English \cite{Ism15,HEA18,HVA20,CHA+21}, Arabic-French 
\cite{AAM18}, Frisian-Dutch \cite{YAK+16}, Mandarin-Taiwanese 
\cite{LLC+06}, Turkish-German \cite{Cet17}, English-Malay \cite{AT12}, English-isiZulu \cite{WN16} and Sepedi-English \cite{MDD13}.
In this work, we build on our ArzEn speech corpus \cite{HVA20}, and enrich it with multiple translations.

\subsection{CSW Machine Translation} While research in CSW MT has been gaining attention over the past years \cite{SK05,DKS18,MMD+19,MLJ+19,SZY+19,TKP21,XY21,CAS+22,HHA+22,GMH+22}, the collected CSW parallel corpora are limited. 
By looking into the reported corpora, we identify a number of dimensions in which they vary. First is  synthetic or human-annotated data. Second, for human-annotated data, it can be either collected, or especially commissioned for MT/NLP.   Third, for collected data, it can be obtained from textual or speech sources. And finally, the data set may include translations to one or more languages. 

To circumvent the data scarcity issue, researchers investigated the use of synthetically generated CSW parallel data for training and testing \cite{GEB20,YHH+20,XY21}. While this is acceptable for training purposes, 
synthetic data should not be used for testing, as it does not reflect real-world CSW distributions. 

For collecting human-annotated parallel corpora, researchers have tried either asking bilingual speakers to translate naturally occurring CSW sentences into monolingual sentences, 
or as another solution to data scarcity, have commissioned annotators to translate monolingual sentences into CSW sentences. 
In Table \ref{table:MT_corpora_rw}, we present a summary of available human-annotated parallel corpora that are focused on CSW. 
For the latter approach, we note that generating CSW data in a human-commissioned fashion could  differ from naturally-occurring CSW sentences. Such data could be biased to the grammatical structure of the monolingual sentences, and could be dominated by single noun switches, 
being the easiest CSW type to generate.

The former approach of translating naturally-occurring CSW sentences into monolingual sentences is the most optimal way to collect a CSW parallel corpus; however, most of the collected corpora rely on CSW sentences obtained from textual sources (mostly from social media platforms). The main concern here is that  CSW phenomena occurring in text are more restricted than those occurring in natural speech. 
In text, people are usually dissuaded from changing scripts, and therefore either avoid switching languages, or switch languages without switching scripts. 
The latter issue was tackled by \newcite{SUH20}, where the authors used a sequence-to-sequence deep learning model to transliterate SMS/chat text collected by \newcite{LDC2017T07} from Arabizi (where Arabic words are written in Roman script) to Arabic orthography. While this corpus is not focused on CSW, it contains CSW sentences. 

Finally, we categorize the collected corpora in terms of the translation direction. 
As shown in Table \ref{table:MT_corpora_rw}, 
most of the corpora include translations for CSW sentences to the secondary language,  
which is most commonly English. A smaller number of researchers investigated translating CSW sentences into the primary language. 
And even fewer researchers included translations to both directions.

The work of \newcite{MLJ+19} is also relevant to our work. The authors extracted Modern Standard Arabic (MSA)-English CSW sentences from the UN documents, to which English translations are available \cite{ECM10}. The Arabic translations were generated by translating the English segments using the Google Translate API. While this can be used for training purposes,  these translations should not be used as gold reference. Moreover, given the nature of the corpus, it contained limited types of CSW, as opposed to the types that occur in conversational speech.

\subsection{CSW Speech Translation} Work on CSW ST is still in its early stages, with little prior work \cite{NKT+19,WSP+22,HUW22}. For CSW ST corpora, two corpora are available for Spanglish: Bangor Miami \cite{CMW04} and Fisher \cite{DDH+14}. 
While the Fisher dataset is not a CSW-focused corpus, it contains a considerable amount of CSW 
\cite{WSP+22}. Similarly, for CSW Egyptian Arabic-English, the Callhome dataset also contains some amount of CSW \cite{GKA+97,KCC+14}. A Japanese-English ST corpus \cite{NKT+19} was also collected, however it includes read-speech and not spontaneous speech.   \newcite{HUW22}  collected a one-hour German-English code-switching speech translation corpus containing read-speech. 

Our new corpus, ArzEn-ST, fills an important resource gap, providing an ST corpus for code-switched Egyptian Arabic-English. The corpus is human-annotated where the source sentences are collected through interviews with bilingual speakers, and contain naturally-generated CSW sentences; they are then translated in both directions: monolingual Egyptian Arabic and monolingual English.

\begin{table*}[ht!]
    \centering
    \setlength{\tabcolsep}{2pt}
    \begin{tabular}{|l|l|}
    \hline
    \multicolumn{1}{|c|}{\textbf{CSW Type}} & \multicolumn{1}{c|}{\textbf{Example}}\\\hline\hline
        \textbf{Inter-sentential CSW}  &  \multicolumn{1}{r|}{It's very difficult making friends at work.   .\<عملت صحاب بس مش فى الشغل>}\\
                                       & \textit{I made friends, but not at work.} {It's very difficult making friends at work.} \\\hline
        \textbf{Extra-sentential CSW}  &
                                        \multicolumn{1}{r|}{ 
                                        \< أنا مولود فى مصر فى القاهرة > 
                                        Okay }\\
                                        & Okay \textit{I was born in Egypt, in Cairo.}\\\hline           
         \textbf{Intra-sentential CSW}  & \multicolumn{1}{r|}{    
                                            related to research \<جزء> actually \<كان فى> 
                                            }\\
                                        & \textit{There was} actually  \textit{a part} related to research\\\hline

          \textbf{Intra-word CSW}      & \multicolumn{1}{r|}{    
                                            \<؟>
                                            \textit{project}+\<لل>
                                            \<بتاعك>
                                            \textit{expectation}+\<ال> 
                                            \<ايه> 
                                        }\\
                                     & \textit{What is your} {expectation} \textit{for the}  {project}?\\\hline\hline

 \textbf{Explicatory CSW} &\multicolumn{1}{r|}{ 
    \<أو مقولة قرتيها فى كتاب وعجبتك؟>
         {quote}
             \<طيب فى مثلا >
        {Okay} 
        }\\
        & 
    {Okay} \textit{\underline{okay} is there a}  quote \textit{\underline{or a quote}} that you read in a book and liked?\\\hline
    
    \textbf{Elaboratory CSW } &\multicolumn{1}{r|}{ 
\<عربي>
 author 
        \<او هي شرقية فهقراها اه من .. من> 
      oriental 
        \<دي يعني هى>
            novel
    \<لو ال>
}\\
&
\textit{If this} novel \textit{is like} oriental \textit{\underline{or it is [originally written] in Arabic} then I will } \\
&  \textit{read it ah from .. from an Arab} author \textit{[in its Arabic version]}.\\\hline

    \end{tabular}
    \caption{Examples of different CSW types in ArzEn followed by their English translation. The originally Arabic phrases are italicized in the English translation. For Explicatory and Elaboratory CSW, the underlining marks the repeated phrases.}
    \label{CSWtypes}
\end{table*}

\section{Overview of the ArzEn Corpus}
\label{sec:arzen}
ArzEn is a conversational speech corpus that is collected through informal interviews. The interviews were held 
at the German University in Cairo, which is a private university where English is the instruction language. The topics discussed were general topics such as education, work and life experiences, career, technology, personal life, hobbies, and travelling experiences. No instructions were given to participants regarding code-switching; they were not asked to produce nor avoid code-switching. Interviews were held with 38 Egyptian Arabic-English bilingual speakers (61.5\% males, 38.5\% females), in the age range of 18-35, who are students (55\%) and employees (45\%) at the university. 
The speech corpus comprises of 12 hours of speech, containing 6,216 sentences.

\subsection{Code-switching Types in ArzEn}
The four main CSW types mentioned in the literature are present in ArzEn \cite{pop80,stefanich2019morphophonology}. We present a corpus example for each of the types in Table~\ref{CSWtypes}.
\paragraph{Inter-sentential CSW} This type of CSW is defined as switching languages from one sentence to another. 

\paragraph{Extra-sentential CSW} This type of CSW, also called \textbf{tag-switching}, is where tag elements from one language are inserted into a monolingual sentence in another language, 
without the need for grammatical considerations. It mostly involves the use of fillers, interjections, tags, and idiomatic expressions. This type of CSW requires only minimal knowledge of the grammar of the secondary language. 

\paragraph{Intra-sentential CSW} This type, also referred to as \textbf{code-mixing}, is defined as using multiple languages within the same sentence, where the CSW segments must conform to the underlying syntactic rules of both languages. 
This type of CSW requires a better understanding of the grammar of both languages, compared to extra-sentential CSW.
 
\paragraph{Intra-word CSW} This type, also called  \textbf{morphological CSW},  is where switching occurs at the level of morphemes. Given that Egyptian Arabic is a morphologically rich language \cite{HEH12}, morphological code-switching occurs  where Egyptians attach Arabic clitics and affixes to English words. 

In addition to the above, and motivated by our interest in translation from CSW texts, we identify two types of repetitive CSW phenomena in terms of their communicative purposes.  

\paragraph{Explicatory CSW} This type of CSW is where the speaker simply repeats the same word in another language.

\paragraph{Elaboratory CSW} This type of CSW is where the speaker code-switches to further elaborate on the  meaning.

Both types are challenging in terms of handling the CSW repetitions when translating into a single language.  We address these issues in   Section~\ref{sec:translation_guidelines}. 

\subsection{Code-switching Statistics in ArzEn}

ArzEn contains a considerable amount of CSW. On the sentence level, 33.2\% of the sentences are monolingual Arabic, 3.1\% are monolingual English, and 63.7\% code-mixed. Among the code-mixed sentences, 46.0\% have morphological CSW. 

On the word level, in the code-mixed sentences, 81.3\% of the words are Arabic, 15.2\% are English, and 3.4\% are morphologically CSW words.  
Morphological CSW in ArzEn involves the use of both Arabic clitics and affixes. A list of the clitics and affixes occurring in morphological code-switched words present in the ArzEn corpus and their frequencies are provided in  \newcite{HDL+22}. 

\subsection{Input Transcription}
The ArzEn collected interviews were manually transcribed by Egyptian Arabic-English bilingual speakers. The transcribers were requested to use Arabic script for Arabic words and Roman script for English words. For morphological CSW words, Arabic clitics and affixes are written in Arabic script and English words are written in Roman script, as follows: \textbf{Arabic prefixes/proclitics + English words \# Arabic suffixes/enclitics}, for example \textit{\<ات>\#TASK+\<ال>}
{\it Al+TASK\#At}\footnote{Transliteration in the HSB scheme \cite{Habash:2007:arabic-transliteration}.} `the+task\#s'. 
While the transcribers generally followed the rules in a strict manner, we observe script confusion 
in the case of borrowed words that have become strongly embedded in  Egyptian Arabic. 
In such cases, transcriptions can contain occurrences of the same words in both scripts, such as \textit{mobile} and \<موبايل>  {\it mwbAyl},  \textit{film} and \<فيلم> {\it fylm}, and \textit{camera} and \<كاميرا> \textit{kAmyrA}.  

Given the spontaneous nature of the corpus, disfluencies were found due to repetitions, corrections, and changing  course/structure mid-sentence. 
Such disfluencies were marked with `..', which occurs in more than 26\% of the corpus sentences. The following tags were also used for non-speech parts: [HES] for hesitation, [HUM] for humming, [COUGH], [LAUGHTER], and [NOISE].

\section{ArzEnST Translation Guidelines}
\label{sec:translation_guidelines}
The transcriptions are translated to monolingual English and monolingual Egyptian Arabic sentences by human translators.\footnote{English translations are performed by one translator, and the dev and test sets are revised by one of the authors. Egyptian Arabic translations are performed by one translator and revised by another.} In this section, we discuss the translation guidelines. 
In general, our decisions are mainly guided by giving a higher priority to fluency over accuracy. We opt for producing as natural as possible outputs that reflect the style of the original sentence. Even though we acknowledge that some of our decisions can make the translation task harder for MT systems, our goal is to produce natural translations. 
The guidelines cover three categories, general translation rules (denoted by \textit{GR}), conversational speech translation rules (denoted by \textit{SR}), and code-switching translation rules (denoted by \textit{CSWR}). In Table \ref{table:guidelines_translation_examples}, we present translation examples covering some of the rules.

\subsection{General Translation Rules (\textit{GR})}
\paragraph{[\textit{GR$_{intended}$}]} Translators are requested to provide natural translations with the intended meaning rather than literal translations. This also covers the case of idiomatic expressions. See Table \ref{table:guidelines_translation_examples}~(a).

\paragraph{[\textit{GR$_{difficult}$}]} Similar to the LDC Arabic-to-English Translation Guidelines \cite{LDC13}, segments that are difficult to translate should be indicated using ((text)). Such cases usually contain highly dialectal Arabic words or Arabic idioms. See Table \ref{table:guidelines_translation_examples}~(b).
\paragraph{[\textit{GR$_{abbrev}$}]} For all abbreviations, we made the decision to provide transliteration as pronounced instead of translation,\footnote{We plan to annotate these cases with full translations in the future, to assist in tasks interested in removing English/Arabic text.} 
for example \textit{NLP} is transliterated as \<ان ال بي> \textit{An Al by}, and \textit{AIESEC} is transliterated as \<آيزيك> {\it {\AMADDA}yzyk}.
\paragraph{[\textit{GR$_{propn}$}]} 
Non-abbreviated proper nouns should be transliterated, unless they have meaning. 
In that case, they should be translated 
as long as the meaning of the sentence remains coherent, otherwise, should be transliterated.\footnote{The translators were advised to refer to Wikipedia Arabic for the translations of titles of books and films.} See Table \ref{table:guidelines_translation_examples}~(c).

\subsection{Conversational Speech Translation Rules (\textit{SR})}
    \paragraph{[\textit{SR$_{style}$}]} Translations should capture the same fluency and style of the original text. This means that disfluencies such as repetitions should also be included in translations. See Table \ref{table:guidelines_translation_examples}~(d).
    \paragraph{[\textit{SR$_{punc+}$}]} Punctuation, non-speech tags, and disfluency marks `..' present in the source text should be kept the same and in the same relative position in the sentence in the target translation.
    \paragraph{[\textit{SR$_{partial}$}]} Due to disfluencies, it is common to have partial Arabic words. 
    We transliterate such partial words, and similar to \newcite{LDC13}, we mark them with a preceding `\%' sign. See Table \ref{table:guidelines_translation_examples}~(e).

\subsection{Code-switching Translation Rules (\textit{CSWR})}

\begin{table*}[th!]
\centering
 \includegraphics[width=\textwidth]{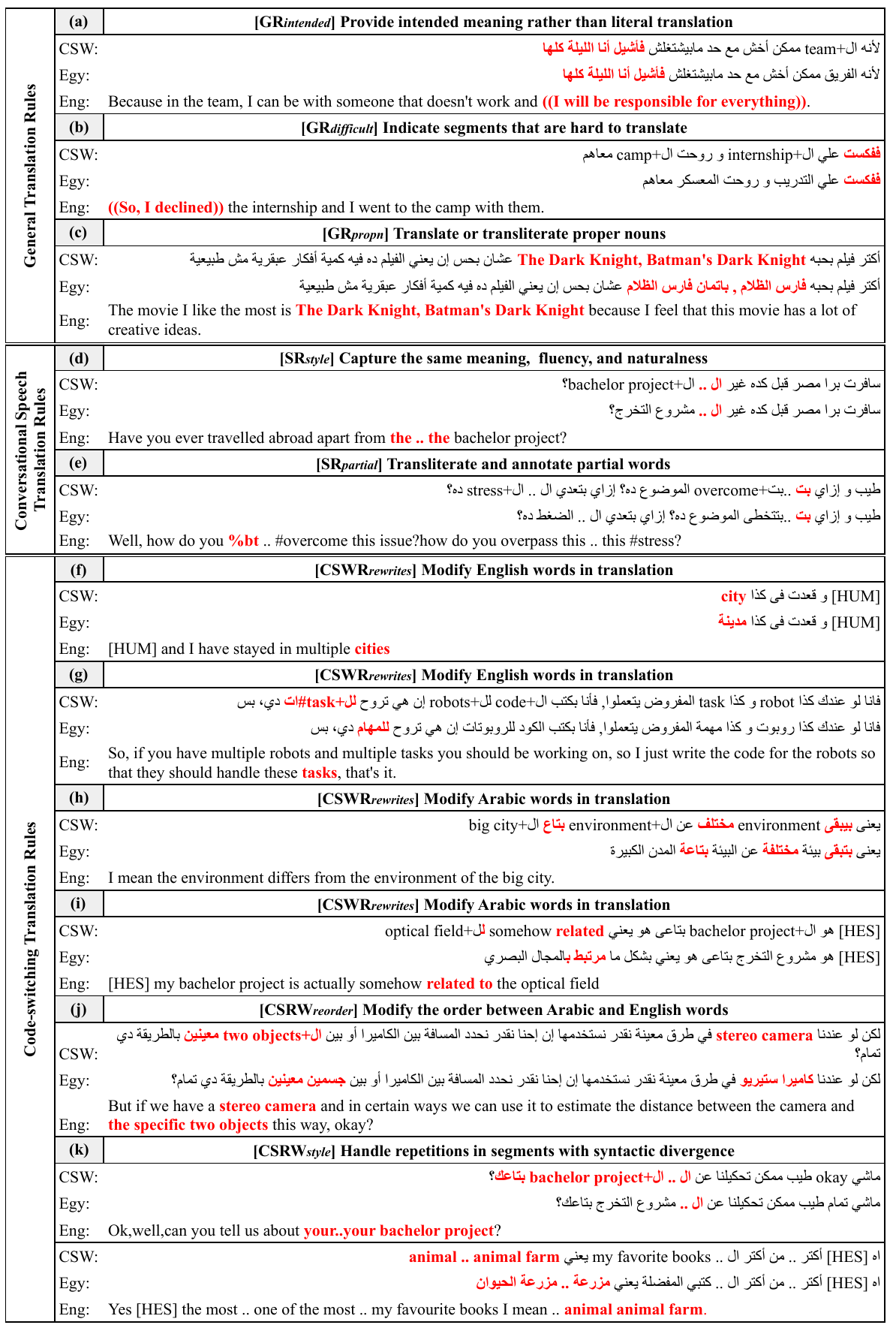}
 \caption{Translation examples following  different guideline rules.} 
 \label{table:guidelines_translation_examples}
\end{table*}

    \paragraph{[\textit{CSWR$_{borrowed}$}]} For English words that are commonly used in Arabic, an attempt should first be made to identify a 
    commonly used 
    reasonable translation, 
    otherwise, translators are allowed to transliterate. Examples of the latter case, included loanwords such as \textit{mobile} and \textit{laptop} that have become strongly integrated in 
    Arabic, as opposed to \textit{online} and \textit{presentation} which can be translated to  \<عبر الأنترنت>  and  \<عرض تقديمي>, respectively.
    \paragraph{[\textit{CSWR$_{rewrites}$}]} 
    We allow modifications to CSW segments when translating into English for the purpose of achieving better fluency. Similarly, when translating into Arabic, we also allow slight modifications to the original Arabic words. We elaborate on such cases for both directions below.
    \paragraph{\textbf{CSW$\rightarrow$En}:} We allow modification to the original English words. This is mainly needed to handle difference in grammatical structures across languages as well as morphological CSW.
    For example, ASK+\<بي> {\it by+ASK}  is translated as `he asks'. See Table \ref{table:guidelines_translation_examples}~(f-g).
    
    \paragraph{\textbf{CSW$\rightarrow$Ar}:} It is allowed to slightly modify the original Arabic words for better fluency. The following are common cases where this is needed.
        \begin{itemize}
            \item Since Arabic makes heavy use of the definite article +\<ال> {\it Al+} to mark different constructions such as adjectival modification and idafa (possessive construct), translators are given permission to drop/reassign the placement of definite articles for the purpose of maximizing fluency. For example, the adjectival construction 
            {\it working life~+\<ال>}
            gets translated to 
            \<الحياة المهنية> (with two definite article instances).
            \item Given that Arabic is a gender-marking language, gender reinflection is sometimes needed to guarantee fluent translation into Arabic.
            See Table \ref{table:guidelines_translation_examples}~(h).
            \item Modifying Arabic prepositions following English words. See Table \ref{table:guidelines_translation_examples}~(i).
        \end{itemize}
    \paragraph{[\textit{CSWR$_{reorder}$}]} We allow changing the order between the original Arabic and English words to handle syntactic divergences between the two languages and achieve better fluency. See Table \ref{table:guidelines_translation_examples}~(j).
    \paragraph{[\textit{CSWR$_{style}$}]} The \textit{SR$_{style}$} rule gets further compounded in the case of CSW when 
    repetition occurs at a location where there is syntactic divergence between both languages, such as adjectival phrases in our language pair. In this case, since the order of words changes during translation, the word to be repeated at that position could also change. Following our fluency preference, we prefer the translation that gives higher fluency over providing an accurate literal translation. See Table \ref{table:guidelines_translation_examples}~(k).
    \paragraph{[\textit{CSWR$_{disfluency}$}]} 
    Another interesting translation challenge arises in the context of CSW when speakers repeat words using different languages. 
    In the case of explicatory CSW, where the English and Arabic words have the exact same meaning, 
    we allow translating the English word into the same present Arabic word, treating it as a case of repetition due to disfluency. In the case of elaboratory CSW, where 
    CSW is used 
    to further elaborate on meaning, we ask the translator to find another translation of the word that would better capture the subtle difference between both words. If that is not possible, we allow the repetition of the word.

\begin{table*}[t]
\centering
\begin{tabular}{|l|r|r|r|r|r|r|}\cline{2-7}
\multicolumn{1}{c|}{}&\multicolumn{2}{c|}{\textbf{ASR}}&\multicolumn{2}{c|}{\textbf{MT}}&\multicolumn{2}{c|}{\textbf{ST}}\\\cline{4-7}
 \multicolumn{1}{c|}{}& \multicolumn{2}{c|}{} &\textbf{CSW$\rightarrow$En} & \textbf{CSW$\rightarrow$Egy} & \textbf{CSW$\rightarrow$En} & \textbf{CSW$\rightarrow$Egy}\\\hline
 \multicolumn{1}{|c|}{\textbf{Training Set}}& \textbf{WER} & \textbf{CER}&\multicolumn{1}{c|}{\textbf{BLEU}} & \multicolumn{1}{c|}{\textbf{BLEU}} & \multicolumn{1}{c|}{\textbf{BLEU}} & \multicolumn{1}{c|}{\textbf{BLEU}}\\\hline\hline
\textbf{ArzEn-ST}  &  57.9   &   36.2 &   8.6  & 48.0  &   4.5  & 13.0\\
\textbf{ArzEn-ST + Extra} &  34.7   &  20.0  &  34.3   &  79.8 &  16.5 &  31.1\\\hline
\end{tabular}
\caption{Summary of results for baseline systems evaluated on ArzEn-ST test set. We present baseline systems for both settings: (1) training using ArzEn-ST data only and (2) training using ArzEn-ST data with \textbf{Extra} monolingual speech corpora for the ASR system and \textbf{Extra} monolingual Egyptian Arabic-English parallel sentences for the MT systems. We report Word Error Rate (WER) and Character Error Rate (CER) for ASR systems, and BLEU score \cite{PRT+02} using SacrebleuBLEU \cite{Pos18} for MT and ST systems.}
\label{table:basline_results}
\end{table*}

\section{Benchmarking Baseline Systems}
\label{sec:baseline_systems}

In this section, we discuss the ASR, MT, and ST baseline systems. We describe the experimental setup for each and present the results in Table \ref{table:basline_results}.
\subsection{Experimental Setup}
We follow the same train, dev, and test splits defined in \newcite{HVA20}. For all the experiments, we use ArzEn-ST dev set (1,402 sentences) for tuning and ArzEn-ST test set (1,470 sentences) for testing. For training, we use ArzEn-ST train set (3,344 sentences), in addition to other monolingual data which we mention below.

\paragraph{Automatic Speech Recognition}
We train a joint CTC/attention based E2E ASR system using ESPnet \cite{WHK+18}. The encoder and decoder consist of 12 and 6 Transformer blocks with 4 heads, feed-forward inner dimension 2048 and attention dimension 256. The CTC/attention weight $(\lambda_1)$ is set to 0.3. SpecAugment \cite{PCZ+19} is applied for data augmentation. For the Language Model (LM), the RNNLM consists of 1~LSTM layer with 1000 hidden units and is trained for 20 epochs. For decoding, the beam size is 20 and the CTC weight is 0.2.

In addition to using ArzEn-ST for training, we also train the ASR system and LM using 
Callhome \cite{GKA+97}, MGB-3 \cite{AVR17}, a 5-hours subset from Librispeech \cite{PCP+15}, and a 5-hours subset from MGB-2 \cite{ABG+16}.\footnote{We followed the setup used in \cite{HDL+22}.} We perform Alif/Ya normalization (Arabic), remove punctuation and corpus-specific annotations, and lower-case English words.\footnote{For the Callhome corpus, we removed partial words.}

\paragraph{Machine Translation}
We train Transformer models using Fairseq \cite{OEB+19} on a single GeForce RTX 3090 GPU. We use the hyperparameters from the FLORES benchmark for low-resource machine translation \cite{GCO+19}. The hyperparameters are given in Appendix~\ref{sec:appendix-MThyperparameters}. For each MT model, we use a BPE model trained jointly on source and target sides. The BPE model is trained using Fairseq with character\_coverage set to $1.0$. We tune the vocabulary size for each experiment for the values of $1k, 3k, 5k, 8k,$ and $16k$. 

In addition to ArzEn-ST, we also train the MT system using $324k$ extra Egyptian Arabic-English parallel sentences obtained from the following parallel corpora: 
Callhome Egyptian Arabic-English Speech Translation
Corpus \cite{KCC+14}, LDC2012T09 \cite{ZMD+12}, 
LDC2017T07 \cite{LDC2017T07}, 
LDC2019T01 \cite{LDC2019T01}, 
LDC2020T05 \cite{LDC2020T05}, and MADAR \cite{BHS18}.\footnote{For corpora with no defined data splits, we follow the guidelines provided in \newcite{DHR+13}.}
%
These extra corpora include $15k$ sentences with CSW instances. 
When translating into En, we use all these extra corpora as Arabic-English training. However,
when translating into Egy, we use these extra corpora as English-Arabic training, but we exclude the $15k$ sentences producing CSW Arabic as our reference does not have CSW sentences.
Data preprocessing involved removing all corpus-specific annotations,  URLs and emoticons, lowercasing, running Moses’ \cite{KHB+07} tokenizer, MADAMIRA \cite{PAD+14} simple tokenization (D0) and  Alif/Ya normalization (Arabic). 

\paragraph{Speech Translation}
We build a cascaded speech translation system, where we train an ASR system and use an MT system to translate the ASR system's outputs. We opt for a cascaded system over an end-to-end system due to the limitation of available resources to build an end-to-end system, in addition to the fact that cascaded systems have been shown to outperform end-to-end systems in low-resource settings \cite{DMV21}.

\subsection{Results}

Table \ref{table:basline_results} presents the results for the MT and ST baseline systems. We also report results for the ASR system used to build the cascaded ST system.\footnote{ASR results are different than those reported in \newcite{HDL+22} as we limit the data to publicly-available corpora, use different preprocessing, and different data splits.} 
We report results for both settings: (1) when training only using ArzEn-ST corpus and (2) when training using ArzEn-ST corpus in addition to the extra monolingual data specified for each task (\textbf{Extra}). As expected, adding extra monolingual data greatly improves results. 
We observe that translating into Arabic achieves higher BLEU scores than translating into English. This is expected, as in the case of translating from CSW text, Arabic words (around 85\% of words) remain mostly the same with possible slight modifications required. 
We also observe that for the ST models, the performance is nearly reduced by half compared to the MT results. This highlights the difficulty of the task. Given that CSW ST has only been slightly tackled by other researchers, we hope that this corpus will motivate further research on this task. 
\section{Conclusion and Future Work}

Code-switching has become a worldwide prevalent phenomenon. This created a need
for NLP systems to be able to handle such mixed input. Code-switched data is typically scarce, which is evident in the limited number of available corpora for machine translation and speech translation tasks. In this paper, we extend the previously collected ArzEn speech corpus with translations to both its primary and secondary languages, providing a three-way code-switched Egyptian Arabic-English speech translation corpus. 
We have discussed the translation guidelines, particularly with regards to issues arising due to the spontaneous nature of the corpus as well as code  switching. We reported benchmark results for baseline ASR, MT, and ST systems. We make this corpus available to motivate and facilitate further research in this area.

For future work, we plan on improving the corpus and using it for code-switching linguistic investigations as well as NLP tasks. With regards to corpus improvements, we plan on adding additional translation references and CODAfying \cite{HDR12,Eskander:2013:processing} the corpus. From a linguistic perspective, having signals from the monolingual Arabic and English translations, we plan to further understand why code-switching occurs at the given points. Finally, we plan to use this corpus for NLP tasks, working on data augmentation for the purpose of improving machine translation and speech translation.

\section*{Acknowledgements}
This project has benefited from financial support by DAAD (German Academic Exchange Service). We also thank the reviewers for their insightful comments and constructive feedback.
\bibliography{mybib}
\bibliographystyle{acl_natbib}

\appendix
\section{MT Hyperparameters}
\label{sec:appendix-MThyperparameters}
The following is the train command:\newline
python3 fairseq\_cli/train.py \${DATA\_DIR} --source-lang src 
--target-lang tgt --arch transformer --share-all-embeddings --encoder-layers 5 --decoder-layers 5 --encoder-embed-dim 512 --decoder-embed-dim 512 --encoder-ffn-embed-dim 2048 --decoder-ffn-embed-dim 2048 --encoder-attention-heads 2 --decoder-attention-heads 2 --encoder-normalize-before --decoder-normalize-before  --dropout 0.4 --attention-dropout 0.2 --relu-dropout 0.2  --weight-decay 0.0001 --label-smoothing 0.2 --criterion label\_smoothed\_cross\_entropy --optimizer adam --adam-betas '(0.9, 0.98)' --clip-norm 0 --lr-scheduler inverse\_sqrt --warmup-updates 4000 --warmup-init-lr 1e-7  --lr 1e-3 --stop-min-lr 1e-9 --max-tokens 4000 --update-freq 4 --max-epoch 100 --save-interval 10 --ddp-backend=no\_c10d

\end{document}